\documentclass[sigconf]{acmart}
\AtBeginDocument{%
  }

\setcopyright{acmlicensed}
\copyrightyear{2027}
\acmYear{2027}
\setcopyright{cc}
\setcctype{by}
\acmConference[WSDM 2027] {Proceedings of the Twentieth ACM International Conference on Web Search and Data Mining}{February 15--19, 2027}{Hong Kong, Hong Kong.}
\acmBooktitle{Proceedings of the Twentieth ACM International Conference on Web Search and Data Mining (WSDM 2027), February 15--19, 2027, Hong Kong, Hong Kong}
\acmISBN{979-8-4007-3117-4/2027/02}
\acmDOI{10.1145/XXXXXX.XXXXXX}

\usepackage{amsmath}
\usepackage{algorithm}
\usepackage{algpseudocode}
\usepackage{booktabs}
\usepackage{multirow}
\usepackage{enumitem}
\usepackage[table]{xcolor}

\begin{document}

\title[Personalized Memory Rubrics]{Personalized Memory Rubrics: Learning What to Remember per User via Rubric-Guided Credit Assignment}

\author{Yupeng Han}
\orcid{0009-0005-7822-0091}
\affiliation{
 \institution{University of Science and Technology of China \& City University of Hong Kong \& State Key Laboratory of Cognitive Intelligence}
 \city{Hefei}
\country{China}
 }
\email{yupenghan@mail.ustc.edu.cn}

\author{Kai Zhang}
\orcid{0000-0001-5335-2470} 
 \authornote{Corresponding authors.}
\affiliation{
 \institution{University of Science and Technology of China \& State Key Laboratory of Cognitive Intelligence }
 \city{Hefei}
\country{China}
 }
\email{kkzhang08@ustc.edu.cn}

\author{Shuochen Liu}
\orcid{0000-0003-4724-8989}
\affiliation{
 \institution{University of Science and Technology of China \& City University of Hong Kong \& State Key Laboratory of Cognitive Intelligence}
 \city{Hefei}
\country{China}
 }
\email{shuochenliu@mail.ustc.edu.cn}

\author{Ze Liu}
\orcid{0009-0008-9664-9504}
\affiliation{
 \institution{University of Science and Technology of China \& State Key Laboratory of Cognitive Intelligence}
 \city{Hefei}
\country{China}
 }
\email{liuze2120@mail.ustc.edu.cn}

\author{Xianquan Wang}
\orcid{0009-0003-9744-220X}
\affiliation{
 \institution{University of Science and Technology of China \& State Key Laboratory of Cognitive Intelligence}
 \city{Hefei}
\country{China}
 }
\email{wxqcn@mail.ustc.edu.cn}

\author{Zhihong Pan}
\orcid{0009-0004-3098-7579}
\affiliation{
 \institution{University of Science and Technology of China \& State Key Laboratory of Cognitive Intelligence}
 \city{Hefei}
\country{China}
 }
\email{panzh@mail.ustc.edu.cn}

\author{Xiangyu Zhao}
\orcid{0000-0003-2926-4416}
 \authornotemark[1]
\affiliation{
 \institution{City University of Hong Kong}
 \city{Hong Kong}
\country{China}
 }
\email{xianzhao@cityu.edu.hk}

\author{Qi Liu}
\orcid{0000-0001-6956-5550}
\affiliation{
 \institution{University of Science and Technology of China \& State Key Laboratory of Cognitive Intelligence}
 \city{Hefei}
\country{China}
 }
\email{qiliuql@ustc.edu.cn}

\author{Enhong Chen}
\orcid{0000-0002-4835-4102}
\affiliation{
 \institution{University of Science and Technology of China \& State Key Laboratory of Cognitive Intelligence}
 \city{Hefei}
\country{China}
 }
\email{cheneh@ustc.edu.cn}

\renewcommand{\shortauthors}{Yupeng Han et al.}

\begin{abstract}
Long-horizon personalized agents maintain an evolving memory by continuously deciding what to add, modify, or discard. However, learning an optimal update policy remains challenging. Conventional training relies solely on terminal question-answering accuracy, yielding a coarse feedback signal that treats all intermediate updates identically. Furthermore, update utility is inherently user-specific, as identical memory actions may yield opposite effects across different user personas. Existing methods fail to address this heterogeneity, relying on globally shared evaluation standards with at most a single scalar intermediate reward. To address these challenges, we propose \textbf{Personalized Memory Rubrics} (\textbf{PMR}), a framework that translates terminal feedback into fine-grained, per-step, and per-user credit assignment. PMR decomposes evaluation into a universal layer capturing cross-persona failure patterns and a per-user adaptive layer modeling individual traits as a specialized delta over the universal baseline. Through \textbf{Rubric-Guided Credit Assignment}, PMR redistributes credit at the token level, selectively amplifying validated decisions on successful trajectories while falling back to standard policy optimization on failed ones to avoid unreliable attribution. The rubric and memory policy \textbf{co-evolve} in a closed loop with anti-circularity guarantees. Crucially, PMR incurs zero inference overhead, operating without any rubric or external judge during deployment. Extensive experiments across four benchmarks demonstrate consistent gains over strong memory-augmented baselines.\footnote{https://github.com/Hyp26cs/WSDM2027\_PMR}
\end{abstract}




\maketitle

\section{Introduction}
\label{sec:intro}

\begin{figure*}
\centering
\includegraphics[width=0.976\linewidth]{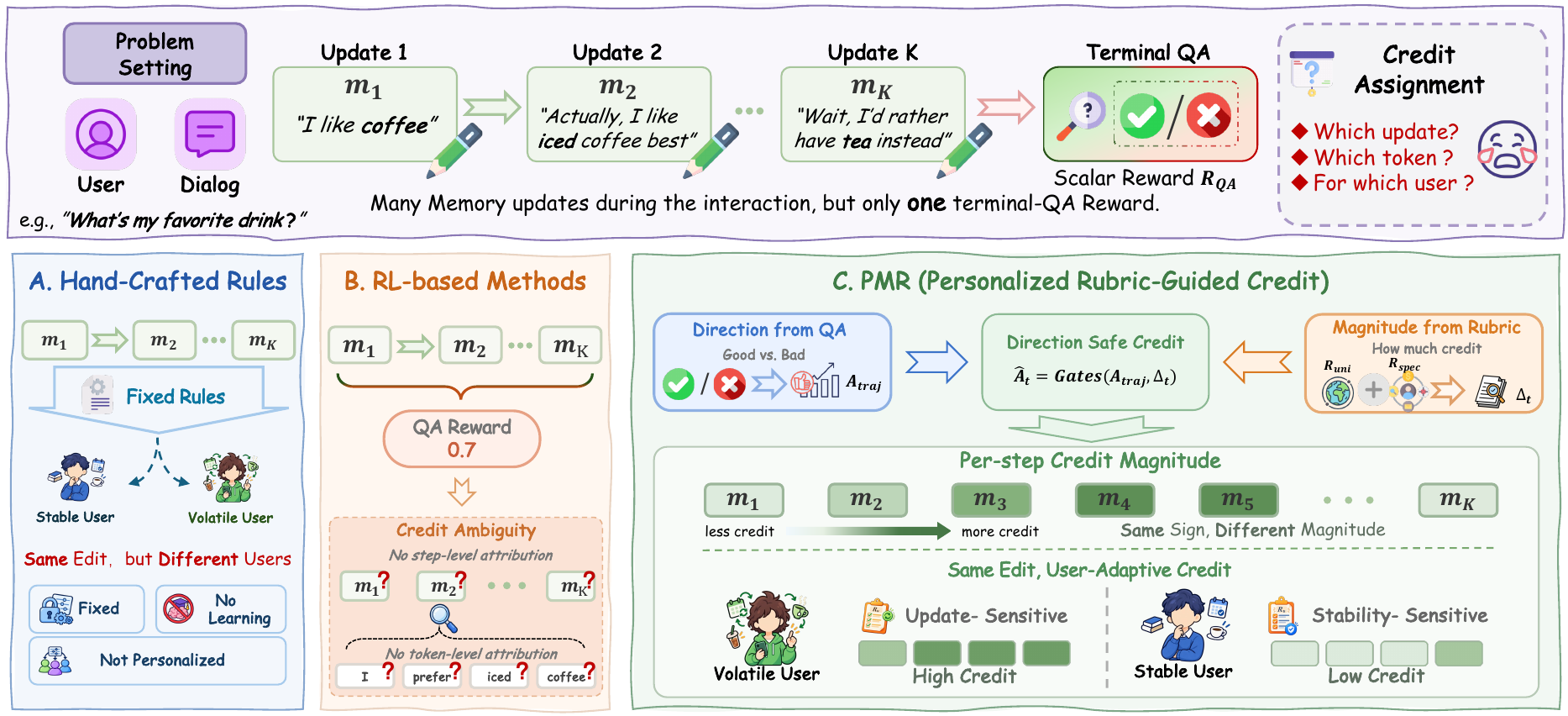}
\vspace{-0.5pt}
\caption{Long-horizon memory learning involves multiple updates but only a terminal QA reward, making fine-grained credit assignment difficult. Hand-crafted methods use fixed, non-personalized rules, while RL-based methods rely on sparse scalar feedback with ambiguous step/token attribution. PMR decouples \textbf{direction} from \textbf{magnitude}: the QA advantage determines the learning direction, while personalized rubric evidence modulates credit magnitude through direction-safe gating.}
\label{fig:motivation}
\end{figure*}


To maintain personalized memory over long horizons, large language model (LLM) agents must continuously distill extensive interactions into reusable representations~\cite{personamem,proac,sam,pcg,aemrl,zhang2025harnessing,rtr}. The cost of processing unbounded histories necessitates localized and sequential memory extraction~\cite{memagent2026,memcoe,hiagent,meta}. As these extractions accumulate, errors in identifying and integrating relevant cues can propagate across updates, compromising the fidelity of the global user model and downstream personalization.


However, mitigating these compounding errors through optimization is typically limited by the credit assignment problem~\cite{2024survey,2025survey,2026survey}. The primary learning signal, terminal question answering (QA) accuracy, is shared across intermediate extractions, obscuring which decisions contributed to or degraded the outcome~\cite{credit,assign,tree}. Early memory systems~\cite{mem0,amem,lightmem,memos2025} address long-horizon memory construction with predefined update rules, avoiding optimization but providing no feedback on extraction quality. More recent reinforcement learning (RL)-based memory agents~\cite{memalpha,agemem,memagent2026,hips} formulate multi-turn memory construction as a Markov Decision Process (MDP)~\cite{huang2026surveyagentmemorysecond, zhang2026landscapeagenticreinforcementlearning}, enabling memory operations to be optimized based on interaction outcomes. However, their rewards remain sparse, leaving the contributions of individual extraction decisions unresolved~\cite{opsd,opdsurvey,rethinkingopd}. Furthermore, dense step-wise rewards, when introduced, measure format compliance or semantic plausibility, providing heuristic signals for surface-level properties rather than downstream utility~\cite{finemem,memt}. MemCoE~\cite{memcoe} incorporates an LLM-judge process reward, but its hand-crafted scoring guide reduces evaluation to a fixed scalar shared across users. Such a uniform criterion is inadequate for personalization, as the utility of a memory extraction is inherently user-specific, where a memory overwrite that benefits a volatile user may prove redundant or harmful for a stable one~\cite{adapaagent,prime,meagent}. This gap motivates a question: \textit{Can sparse, terminal feedback be transformed into fine-grained, per-user credit assignment without relying on an external, hand-crafted judge?}

To this end, we introduce \textbf{Personalized Memory Rubrics} (\textbf{PMR}), a framework that bridges terminal feedback to per-step, per-user credit assignment within Group Relative Policy Optimization (GRPO)~\cite{grpo}. As illustrated in Fig.~\ref{fig:motivation}, PMR decomposes the evaluation rubric into a universal component and a user-adaptive component through three mechanisms: (1)~\textbf{Rubric Induction} learns a diagnostic rubric for each user by contrasting successful and failed rollouts across diverse memory quality dimensions. The universal component captures failure patterns that generalize across users through cross-user comparisons, while the user-adaptive component models residual failure modes specific to each user on top of the universal layer. Before injection, the rubric's discriminative power is verified against real outcomes to ensure it provides a meaningful signal. (2)~\textbf{Rubric-Guided Credit Assignment} uses the induced rubric as a privileged context for a frozen teacher model, which evaluates the student's sampled trajectories and produces per-token endorsement scores. These scores modulate the magnitude of each token's GRPO advantage through a direction gate (amplify-only, preventing the rubric from reversing the outcome direction) and an entropy gate (focusing credit on edit decisions rather than carry-over tokens). On successful trajectories, endorsed tokens receive amplified credit; on failed trajectories, standard GRPO is retained, as the rubric is more reliable for identifying key contributing tokens than for attributing failure to specific decisions. (3)~The \textbf{Co-Evolution Loop} alternates between rubric induction and credit assignment, allowing the rubric to guide credit allocation while student rollouts provide feedback for its refinement. To avoid circular dependencies, the rubric modulates only the \emph{magnitude} of credit, while the \emph{direction} remains determined by the trajectory-level QA outcome. Unlike prior approaches that require an external judge at inference time~\cite{memcoe}, PMR is fully self-contained and requires no rubric during inference. Our main contributions are summarized as follows:

\begin{itemize}[leftmargin=*, topsep=1pt, itemsep=0pt]
    \item We formulate the \textbf{per-user credit assignment} problem in long-horizon personalized memory and propose \textbf{PMR}, a framework that decomposes rubrics into universal and user-adaptive layers to translate sparse terminal QA rewards into fine-grained, step-wise feedback within GRPO.
    \item We introduce \textbf{rubric-guided credit assignment} with an asymmetric design: the rubric amplifies endorsed decisions on successful trajectories while preserving standard GRPO on failed ones, supported by direction and entropy gates that bound the rubric's influence and focus credit on edit decisions.
    \item Extensive experiments on PersonaMem, PrefEval, PersonaBench and PERMA demonstrate consistent improvements. Ablations isolate each design choice, including the asymmetric design, per-user personalization, and co-evolution.
\end{itemize}

\section{Related Work}
\label{sec:related}

\begin{figure*}[t]
\centering
\includegraphics[width=\linewidth]{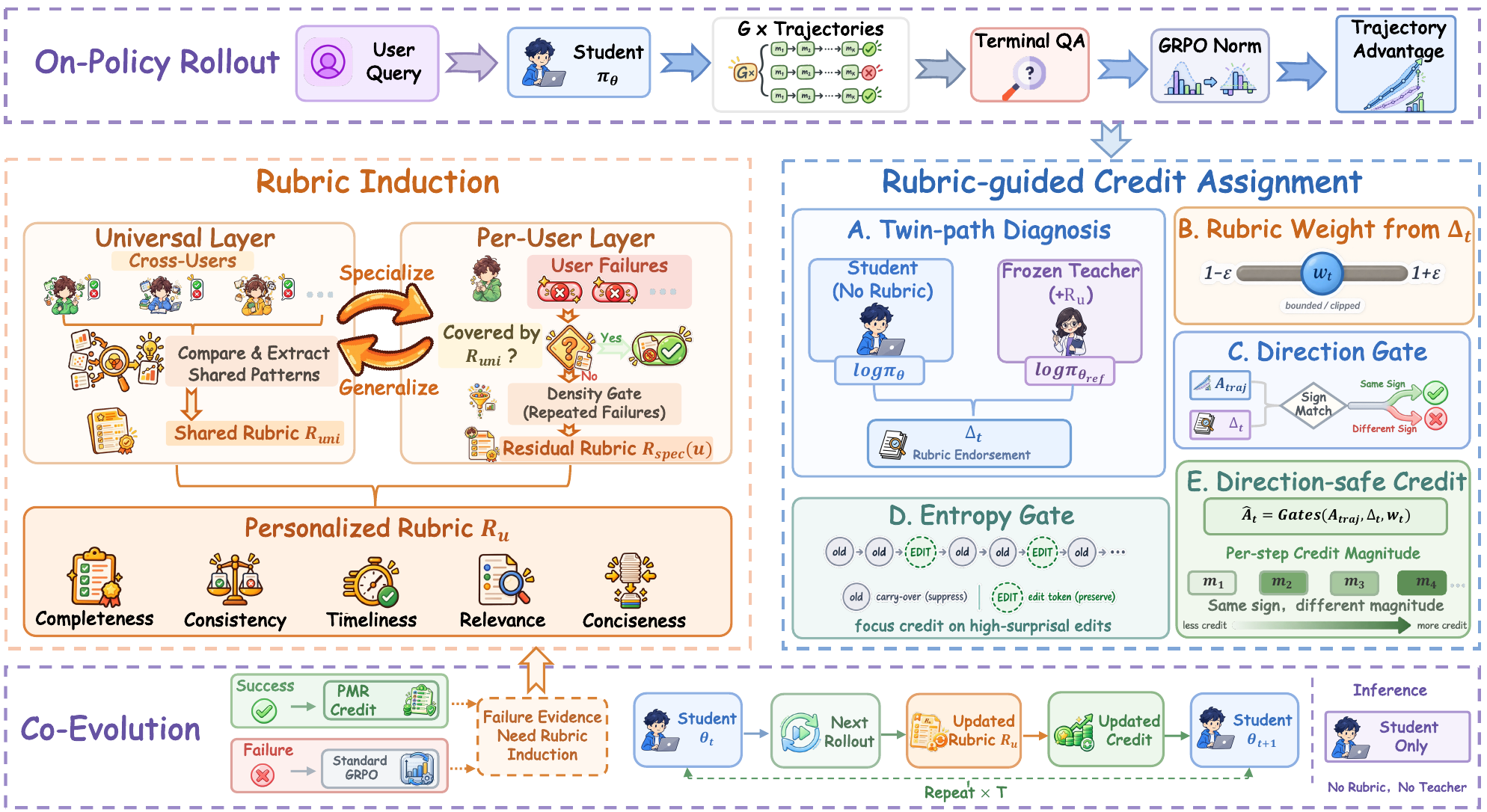}
\caption{\textbf{Overview of PMR.}
PMR first induces a personalized rubric $R_u$ from shared cross-user patterns and residual per-user failures, and then uses it to shape direction-safe, per-step credit while preserving the trajectory-level update direction.
A co-evolution loop iteratively updates the student, rubric, and credit assignment, with standard GRPO used as a fallback for failures.
At inference, only the student model is required.}
\label{fig:method}
\end{figure*}

\subsection{Memory-Augmented LLM Agents}
\label{sec:related-memory}

Early memory systems~\cite{memgpt,amem,lightmem,mem0,rfmem} rely primarily on predefined or model-driven procedures for memory extraction, organization, and consolidation, leaving update policies fixed rather than optimized from downstream outcomes. CoM~\cite{xu2026chainofmemorylightweightmemoryconstruction} introduces an iterative reasoning mechanism that structures historical synthesis through chained memory contexts. More recent RL-based methods~\cite{memskill,kang2026memreaderpassiveactiveextraction,memo,trimem} formulate memory construction as a sequential decision process and optimize memory operations using interaction-level feedback. MemAgent~\cite{memagent2026} processes long contexts and learns memory updates through RL, while Mem-$\alpha$~\cite{memalpha} directly derives rewards from downstream performance. MemBuilder~\cite{membuilder} attributes dense rewards to individual memory components based on their downstream contribution, introducing contribution-aware gradient weighting. However, these methods either rely on terminal rewards, attribute credit at the memory component level rather than the token level, or compress evaluation into scalar scores shared across users. MemCoE~\cite{memcoe} employs an intermediate LLM judge to evaluate memory updates, yet its scoring guide is hand-crafted, scalar, and globally identical. This is limiting for personalized memory, where the utility of an extraction depends on the user's evolving preferences and history~\cite{wu2026knowmebenchbenchmarkingpersonunderstanding}. Although personalized agents adapt to users~\cite{adapaagent,meagent,dppm}, they treat profiles as static context rather than dynamic reward criteria, precluding fine-grained, user-tailored credit assignment. PMR addresses this gap by learning a user-adaptive evaluation rubric that redistributes token-level credit within GRPO, shifting the focus from learning only \emph{what to do} to also learning \emph{what to evaluate}.

\subsection{From Trajectory to Token-Level Supervision}
\label{sec:related-opsd}

On-policy self-distillation (OPSD)~\cite{opsd} introduces a teacher-student framework in which a frozen policy augmented with privileged context, such as reference answers, provides per-token guidance on the student's sampled trajectories. The teacher's next-token distribution serves as a dense supervision signal through KL divergence, providing finer-grained credit information than conventional RL, which assigns a trajectory-level reward uniformly across tokens. RLSD~\cite{rlsd} extends this framework by combining RL policy gradients with self-distillation, using trajectory-level advantages to determine the optimization direction and token-level teacher-student divergence to modulate its magnitude. RCSD~\cite{rcsd} replaces reference-answer supervision with a learned rubric as privileged context, demonstrating the value of structured evaluation criteria for fine-grained supervision, while PARL~\cite{parl} further explores rubric induction from interaction histories. However, these approaches hinge on the quality of the privileged supervision. When the privileged context is provided by a learned and potentially noisy rubric, direct distribution matching may propagate erroneous guidance into the policy. CREST~\cite{crest} addresses this issue with gating mechanisms, including an amplify-only direction gate and entropy-based edit focus, which constrain privileged supervision to credit magnitude while preserving the direction determined by the verifier. However, its direction signal relies on environment-provided per-turn rewards, which are unavailable in terminal-only settings. Our work addresses this limitation through \emph{credit magnitude modulation}: a learned per-user rubric modulates the magnitude of each token's advantage within GRPO, while its direction is determined exclusively by the terminal QA outcome. This design retains the fine-grained supervision of OPSD-style methods without directly matching the policy to potentially noisy rubric distributions, while the co-evolving per-user rubric enables evaluation criteria to adapt to both user-specific characteristics and policy improvement.

\section{Method}
\label{sec:method}

In this section, we present \textbf{Personalized Memory Rubrics} (\textbf{PMR}), a framework that bridges terminal QA feedback to per-step, per-user credit assignment. \emph{Rubric Induction} (Sec.~\ref{sec:method-induction}) learns a per-user rubric from success/failure rollout comparison, decomposed into a universal layer and a per-user layer. \emph{Rubric-Guided Credit Assignment} (Sec.~\ref{sec:method-credit}) uses the learned rubric to redistribute credit at the token level, amplifying endorsed decisions on successful trajectories while preserving standard policy optimization on failed ones. The two stages \emph{co-evolve} in a closed loop (Sec.~\ref{sec:method-coop}). The overall architecture is illustrated in Fig.~\ref{fig:method}.

\subsection{Preliminary}
\label{sec:method-prelim}

We consider a conversational setting in which a memory-augmented agent interacts with user $u$ across sessions. The dialogue history is segmented into contextual chunks $\{c_1, \ldots, c_K\}$, which are processed sequentially. At each memory-update step $k \in \{1, \ldots, K\}$, the agent maintains a compact memory state $m_k$ (a natural-language summary), which is updated as:
\begin{equation}
    m_k = \pi_\theta(c_k, m_{k-1}),
    \label{eq:memory-update}
\end{equation}
where $\pi_\theta$ is the memory update policy. After processing all $K$ chunks, the agent answers a question $q$ using $m_K$. The correctness reward $R_{\text{ans}} \in \{0, 1\}$ is the only observable feedback signal.

We adopt Group Relative Policy Optimization (GRPO)~\cite{grpo} as the base training algorithm. For each prompt, $G$ trajectories are sampled; the trajectory-level advantage is computed via group-normalized rewards:
\begin{equation}
A^{\text{traj}}_i = \frac{R_i - \bar{R}}{\sigma_R + \varepsilon},
\label{eq:grpo}
\end{equation}
where $R_i$ is the reward of trajectory $i$, $\bar{R}$ and $\sigma_R$ are the group mean and standard deviation, respectively, and $\varepsilon>0$ is a small constant for numerical stability. In standard GRPO, every token in trajectory $i$ receives the same $A^{\text{traj}}_i$, meaning all update decisions share a single trajectory-level signal.

\paragraph{\textbf{The credit assignment problem.}}
Since $R_{\text{ans}}$ is computed only at the end, all $K$ updates share one reward. The agent cannot determine whether the third update (a good preference overwrite) or the seventh (a stale-detail retention) contributed to the outcome. Prior RL-based memory agents~\cite{memalpha,memagent2026} face the same limitation: all $K$ updates within an episode share a single reward signal, and their dense per-step signals (format compliance, semantic plausibility) evaluate surface properties rather than downstream utility. MemCoE~\cite{memcoe} mitigates this with an intermediate LLM-judge process reward, but its scoring guide is hand-crafted, collapses to a scalar, and is globally identical for every user.

\paragraph{\textbf{The personalization challenge.}}
The value of an update is inherently user-specific: the same overwrite is beneficial for a volatile user yet harmful for a stable one. Effective credit assignment must be both \emph{per-step} and \emph{per-user}. To enable this, we decompose the evaluation rubric into shared and personalized components:
\begin{equation}
    \mathcal{R}_u = \mathcal{R}_{\text{uni}} \cup \mathcal{R}_{\text{spec}}(u),
    \label{eq:rubric-decomp}
\end{equation}
where $\mathcal{R}_{\text{uni}}$ defines general evaluation criteria shared across users, while $\mathcal{R}_{\text{spec}}(u)$ captures user-specific criteria that reflect the behavioral patterns and preferences of user $u$.






\subsection{Rubric Induction}
\label{sec:method-induction}

Existing evaluation criteria for memory updates are either \emph{hand-crafted} (e.g., MemCoE's fixed rubric-following scores~\cite{memcoe}) or absent entirely (standard GRPO provides no per-step evaluation signal~\cite{grpo}). Those that exist are \emph{globally shared} across users and \emph{frozen} before training begins. This is inadequate for two reasons: a static, global criterion cannot capture user-specific failure patterns, and a frozen criterion cannot adapt to new failure modes that emerge as the policy evolves. To address these limitations, we propose learning a \emph{diagnostic rubric} from the agent's own rollout experience. The rubric is a natural-language evaluation document organized around five memory quality dimensions: \emph{completeness}, \emph{timeliness}, \emph{consistency}, \emph{conciseness}, and \emph{relevance}. For each dimension, it specifies a universal goal that defines what to evaluate, a user-specific standard that determines how strictly to evaluate it, and user-specific diagnostic checks that identify concrete questions for verification. The complete rubric for user $u$ is $\mathcal{R}_u = \mathcal{R}_{\text{uni}} \cup \mathcal{R}_{\text{spec}}(u)$, where $\mathcal{R}_{\text{uni}}$ contains universally shared criteria and $\mathcal{R}_{\text{spec}}(u)$ contains per-user adaptations. Learning such a rubric presents two challenges: the rubric has \emph{no ground truth} to supervise against, and per-user rollout data is \emph{sparse}. Our approach addresses both by \emph{pooling evidence across users} to learn the universal layer and \emph{verifying the discriminative power} of the resulting rubric before using it for evaluation.

\paragraph{\textbf{Universal layer from cross-persona comparison.}}
The universal layer $\mathcal{R}_{\text{uni}}$ is learned by comparing successful and failed rollouts across all training personas. We collect success/failure pairs, preferring same-persona pairs for fairer comparison and supplementing with cross-persona pairs to identify universal patterns. Multiple pairs from different users are presented to an LLM, which identifies failure patterns that are \emph{universal} (appearing across different users) versus \emph{per-user only}. Only universal patterns are added to $\mathcal{R}_{\text{uni}}$; per-user patterns are deferred to $\mathcal{R}_{\text{spec}}$.

\paragraph{\textbf{Per-user layer as residual.}}
The per-user layer $\mathcal{R}_{\text{spec}}(u)$ is learned as a \emph{residual} on top of $\mathcal{R}_{\text{uni}}$. The induction prompt injects $\mathcal{R}_{\text{uni}}$ as known context and asks: \textit{Given these universal rules, what patterns does this user need that are not already covered?}. A coverage test determines whether per-user adaptation is needed: if $\mathcal{R}_{\text{uni}}$ already covers the majority of this user's failure patterns, $\mathcal{R}_{\text{spec}}(u)$ is allowed to be empty. Within per-user induction, a failure density gate marks a dimension as active only if its failure rate exceeds $\theta_{\text{active}}$, ensuring only dimensions with repeated failures receive adjustments.

\paragraph{\textbf{Quality gate and cross-persona sharing.}}
Before injecting the rubric into the teacher's prompt for credit assignment (Sec.~\ref{sec:method-credit}), we verify its discriminative power by computing $\mathrm{corr}(\bar{\Delta}_i, A^{\text{traj}}_i)$ across all trajectories, where $\bar{\Delta}_i$ is the mean endorsement of trajectory $i$. If the correlation falls below a threshold, the rubric lacks signal, and the round reverts to standard GRPO. Additionally, if a per-user check appears in $\geq \theta_{\text{flow}}$ fraction of personas, it is promoted to $\mathcal{R}_{\text{uni}}$ and pruned from individual $\mathcal{R}_{\text{spec}}$ layers, as it represents a universal pattern rather than a localized adaptation.

\subsection{Rubric-Guided Credit Assignment}
\label{sec:method-credit}

Standard GRPO assigns a single trajectory-level advantage $A^{\text{traj}}$ uniformly to every token in a trajectory, providing no per-token resolution to distinguish which update decisions contributed to the outcome. Prior attempts to densify this signal remain insufficient: heuristic per-step rewards such as format compliance or semantic plausibility evaluate \emph{surface properties} rather than downstream utility, while scalar intermediate rewards~\cite{memcoe} reduce memory quality to a single value without indicating which dimension failed or what to fix. A more principled approach, suggested by the privileged information framework (Sec.~\ref{sec:related-opsd}), is to use the learned rubric as the teacher's privileged context. However, applying KL distillation in this setting \emph{forces the student to match the teacher's full output distribution}: when the privileged context (a learned, inherently noisy rubric) is wrong, the entire policy is pulled toward incorrect distributions, causing training collapse. We instead propose \emph{credit magnitude modulation}, which separates \emph{direction} (always from real QA outcomes) from \emph{magnitude} (modulated by the rubric). To distinguish the two levels of granularity, we use $k$ to index memory-update steps and $t$ to index token positions within an update. For the $k$-th memory update, let $y_{k,t}$ denote its $t$-th generated token; when the memory-update index is clear from context, we omit $k$ for notational simplicity. The modified advantage for token $t$ is:
\begin{equation}
\hat{A}_t = A^{\text{traj}} \cdot \Big[1 + \lambda \cdot g^{\text{dir}}_t \cdot m^{\text{ent}}_t \cdot (w_t - 1)\Big],
\label{eq:master}
\end{equation}
where three independent factors modulate the magnitude: the rubric weight $w_t$, the direction gate $g^{\text{dir}}_t$, and the entropy gate $m^{\text{ent}}_t$. The trajectory-level direction $A^{\text{traj}}$ is never modified. The coefficient $\lambda$ controls the overall rubric influence, with a linear warmup over the first few steps to align with rubric cold-start.

\paragraph{\textbf{Rubric endorsement and bounded weight.}}
The rubric is injected into the teacher's prompt as privileged context. The teacher is the same model with frozen weights $\pi_{\theta_{\text{ref}}}$; the student is the trainable $\pi_\theta$. Both evaluate the student's own sampled trajectory (on-policy, no gradient for the teacher). The per-token rubric endorsement is:
\begin{equation}
    \Delta_t = \beta\Big[\log\pi_{\theta_{\text{ref}}}(y_t \mid x \oplus \mathcal{R}_u, y_{<t}) - \log\pi_\theta(y_t \mid x, y_{<t})\Big],
    \label{eq:delta}
\end{equation}
where $\beta>0$ is a scaling coefficient that controls the magnitude of rubric endorsement and $x \oplus \mathcal{R}_u$ denotes the prompt with the user's rubric prepended. A positive $\Delta_t$ indicates the rubric increases the teacher's probability of token $y_t$ (\emph{endorsement}); a negative $\Delta_t$ indicates the opposite. The bounded weight is:
\begin{equation}
    w_t = \mathrm{clip}\Big(e^{\,\Delta_t},\; 1 - \epsilon,\; 1 + \epsilon\Big),
    \label{eq:weight}
\end{equation}
where $\epsilon \in (0,1)$ bounds the per-token adjustment to $[1-\epsilon, 1+\epsilon]$. Endorsed tokens ($\Delta_t > 0$) receive $w_t > 1$ and amplified credit; for $\Delta_t < 0$ the direction gate below closes, leaving $\hat{A}_t = A^{\mathrm{traj}}$, so non-endorsement withholds amplification rather than penalizing. The clip ensures no single token dominates the gradient.

\paragraph{\textbf{Direction gate.}}
The direction gate ensures the rubric can only \emph{amplify} existing credit, never \emph{reverse} it:
\begin{equation}
    g^{\text{dir}}_t = \mathbb{1}\Big[\mathrm{sign}(A^{\text{traj}}) \cdot \Delta_t > 0\Big].
    \label{eq:direction-gate}
\end{equation}
When the endorsement direction disagrees with the trajectory-level outcome, the gate closes and the modulation vanishes. Without it, such evidence would shrink a token's credit to $(1-\lambda\epsilon)A^{\text{traj}}$, penalizing it on an otherwise successful trajectory; with it, the worst case is imprecise amplification, never wrong-direction learning.

\paragraph{\textbf{Entropy gate.}}
Memory updates comprise two token types: \emph{carry-over} tokens copied from the previous memory (low surprisal) and \emph{edit} tokens representing new decisions (high surprisal). Carry-over tokens are not decisions and should not bear credit. We use the student's per-token surprisal $s_t = -\log\pi_\theta(y_t \mid x, y_{<t})$ to distinguish them:
\begin{equation}
m^{\text{ent}}_t =
\operatorname{sigmoid}\Big(
\frac{s_t - \mu_s}{\sigma_s + \varepsilon}
\Big),
\end{equation}
where $\mu_s$ and $\sigma_s$ denote the mean and standard deviation of surprisal across the trajectory, respectively, and $\varepsilon>0$ is the same small numerical-stability constant used in the GRPO reward normalization. Carry-over tokens receive $m^{\text{ent}}_t \approx 0$ (credit suppressed), whereas edit tokens receive $m^{\text{ent}}_t \approx 1$ (credit preserved). This incurs no additional forward-pass cost, since $s_t$ is already available from the student forward pass.

\paragraph{\textbf{Asymmetric design.}}
The full credit assignment mechanism is applied only on \emph{successful} trajectories ($A^{\text{traj}} > 0$); \emph{failed} trajectories ($A^{\text{traj}} \leq 0$) receive standard GRPO. On successful trajectories, the rubric performs \emph{credit sorting}: given the trajectory succeeded, which tokens were key? This is a ranking task tolerant to rubric noise, as the worst case is imprecise amplification. On failed trajectories, the rubric would need to perform \emph{blame attribution}: given the trajectory failed, which tokens caused the failure? This causal task is fragile, as rubric misattribution can penalize correct tokens and teach wrong lessons. Failed trajectories' evidence is not discarded but feeds the rubric induction pipeline (Sec.~\ref{sec:method-induction}), which learns from failure patterns without gradient risk.

\paragraph{\textbf{Safety guarantees.}}

The modulation satisfies three properties that limit rubric errors. First, the modulation factor is bounded below by $1 - \lambda\epsilon > 0$, so the rubric scales credit magnitude without flipping its sign. Second, $\epsilon$ bounds the per-token adjustment, preventing excessive credit concentration. Third, when $A^{\text{traj}} = 0$ or $\Delta_t \approx 0$, the modified advantage reduces to standard GRPO.

\subsection{Co-Evolution Loop}
\label{sec:method-coop}

A rubric learned from the rollouts of an initially weak student captures only the failure patterns of that policy. As the student improves, new failure modes emerge while old ones disappear, rendering a \emph{frozen} rubric increasingly stale. We address this issue through a \emph{co-evolution loop} that alternates rubric induction (Sec.~\ref{sec:method-induction}) and credit assignment (Sec.~\ref{sec:method-credit}). In each round, the student generates new trajectories, the rubric is refined by comparing success and failure patterns, and credit is redistributed using the updated rubric. Repeating this process allows the rubric to track the student's evolving failure landscape, while the resulting credit signals continuously guide policy improvement.

\paragraph{\textbf{Anti-circularity.}}
A potential concern is that the rubric, which guides credit allocation, is itself learned from the student's rollouts, creating a circular dependency. Our design offers a \emph{stronger} guarantee through the structural separation in Eq.~\ref{eq:master}: the rubric operates exclusively in the \emph{magnitude channel} ($w_t$), while \emph{direction} ($A^{\text{traj}}$) is always determined by real QA outcomes. The rubric cannot create a positive signal for a bad trajectory or a negative signal for a good one. It can only redistribute credit among tokens within a trajectory whose direction is already fixed. This is not merely isolating two signals but \emph{structurally confining} the rubric's influence to a channel that cannot affect gradient direction.

\paragraph{\textbf{Inference.}}
At inference time, the student operates independently: it generates memory updates without any rubric context, as the rubric's diagnostic knowledge has been internalized through credit-guided training. This makes PMR fully \emph{self-contained}, unlike MemCoE~\cite{memcoe} which requires an external strong LLM judge at both training and inference time.

\section{Experiments}
\label{sec:exp}

\subsection{Experimental Setup}

\paragraph{\textbf{Benchmarks.}}
As described in Tab.~\ref{tab:benchmarks}, we evaluate on four personalized memory benchmarks. \textbf{PersonaMem}~\cite{personamem} simulates multi-session dialogues for 20 personas whose preferences evolve across 15 task domains, with 7 query types probing distinct personalization capabilities including fact recall, preference tracking, and cross-domain generalization. We use context scales of 32K and 128K tokens. \textbf{PrefEval}~\cite{prefeval} evaluates explicit and implicit preference following with 50 inserted distractor turns between preference revelation and query, testing long-range retrieval under noise. \textbf{PersonaBench}~\cite{personabench} assesses personalized retrieval over heterogeneous user corpora (conversations, AI dialogues, e-commerce histories) across four noise levels (0, 0.3, 0.5, 0.7). \textbf{PERMA}~\cite{perma} evaluates cross-domain generalization with 10 user profiles across 20 domains, structured as Clean/Noise $\times$ Single/Multi-domain scenarios. PersonaMem serves as the in-domain benchmark; the remaining three are out-of-domain. We report \textbf{accuracy} for PersonaMem, PrefEval, and PERMA, and \textbf{macro-F1} for PersonaBench.

\begin{table}[t]
\centering
\small
\setlength{\tabcolsep}{3pt}{%
\begin{tabular}{lccccc}
\toprule
\textbf{Benchmark} & \textbf{Type} & \textbf{\#Users} & \textbf{\#Queries} & \textbf{Settings} & \textbf{Metric} \\
\midrule
PersonaMem & ID & 20 & 589 / 2727 & 32K, 128K & Accuracy \\
PrefEval   & OOD       & -- & 1000 / 1000 & Explicit, Implicit & Accuracy \\
PersonaBench & OOD     & 6  & 263  & 4 noise levels & Macro-F1 \\
PERMA      & OOD       & 10 & 705 / 389  & C/N $\times$ S/M & Accuracy \\
\bottomrule
\end{tabular}%
}
\caption{Summary of evaluation benchmarks. PersonaMem is in-domain; the rest are out-of-domain.}
\label{tab:benchmarks}
\vspace{-0.5cm}
\end{table}

\paragraph{\textbf{Baselines}}

We compare against a diverse selection of baselines. \textbf{Long Context} feeds raw dialogue histories directly into the context window, serving as a memoryless lower bound, while \textbf{RAG} retrieves the top-$k$ most relevant segments from the full history via dense retrieval. We include three static memory pipelines that maintain external memory banks with fixed rules: \textbf{Mem0}~\cite{mem0} applies hand-crafted extraction, summarization, and selective forgetting; \textbf{A-Mem}~\cite{amem} organizes memory as a self-structuring Zettelkasten with dynamic connections; and \textbf{LightMem}~\cite{lightmem} maintains memory within a fixed token budget via lightweight compression. We further compare against three RL-based memory agents that learn memory operations from interaction outcomes: \textbf{MemAgent}~\cite{memagent2026} reshapes long-context interactions via multi-conversation RL with terminal QA reward; \textbf{Mem-$\alpha$}~\cite{memalpha} learns memory construction via RL with composite rewards (correctness, format, compression); and \textbf{MemSkill}~\cite{memskill} reframes static memory operations as learnable and evolvable skills with dynamic selection and refinement. Finally, we include the strategy-guided method: \textbf{MemCoE}~\cite{memcoe} introduces a two-stage optimization with guideline induction via textual gradient and GRPO with an intermediate LLM judge reward.
\begin{table*}[!htbp]
\centering
\resizebox{\textwidth}{!}{%
\begin{tabular}{l|cc|cc|cccc|cccc}
\toprule
\multirow{2}{*}{\textbf{Method}} & \multicolumn{2}{c|}{\cellcolor{red!12}\textbf{PersonaMem}} & \multicolumn{2}{c|}{\cellcolor{blue!12}\textbf{PrefEval}} & \multicolumn{4}{c|}{\cellcolor{orange!12}\textbf{PersonaBench}} & \multicolumn{4}{c}{\cellcolor{teal!12}\textbf{PERMA}} \\
& \cellcolor{red!12}32K & \cellcolor{red!12}128K & \cellcolor{blue!12}Expl. & \cellcolor{blue!12}Impl. & \cellcolor{orange!12}0 & \cellcolor{orange!12}0.3 & \cellcolor{orange!12}0.5 & \cellcolor{orange!12}0.7 & \cellcolor{teal!12}C-S & \cellcolor{teal!12}C-M & \cellcolor{teal!12}N-S & \cellcolor{teal!12}N-M \\
\midrule
Long Context & 45.18 & 35.57 & 34.70 & 27.30 & 27.77 & 22.38 & 19.45 & 12.59 & 24.11 & 22.88 & 24.82 & 22.81 \\
RAG          & 54.22 & 38.50 & 47.40 & 32.40 & 32.21 & 29.13 & 25.28 & 23.36 & 51.63 & 36.84 & 51.77 & 35.09 \\
\midrule
Mem0         & 48.53 & 39.67 & 59.10 & 48.80 & 17.43 & 18.95 & 19.31 & 16.49 & 52.76 & 51.67 & 51.49 & 51.15 \\
A-Mem        & 59.63 & 40.34 & 64.70 & 54.30 & 30.09 & 28.52 & 25.81 & 24.10 & 53.05 & 52.70 & 52.91 & 49.87 \\
LightMem     & 54.22 & 37.40 & 67.20 & 58.90 & 19.14 & 17.83 & 19.61 & 17.51 & 59.43 & 53.73 & 59.43 & 55.53 \\
\midrule
MemAgent     & 61.44 & 51.34 & 77.40 & 64.40 & 30.74 & 28.34 & 24.47 & 23.53 & 53.98 & 53.90 & 55.74 & 48.84 \\
Mem-$\alpha$ & 59.63 & 49.25 & 75.70 & 63.10 & 19.31 & 18.01 & 17.35 & 16.64 & 52.19 & 51.67 & 54.46 & 46.53 \\
MemSkill     & 66.27 & 55.74 & 82.60 & 68.40 & 31.19 & 29.89 & 24.85 & 23.36 & 62.83 & 51.92 & 60.14 & 53.21 \\
MemCoE       & 67.47 & 58.67 & 83.40 & 68.60 & 31.71 & 29.03 & 25.12 & 23.74 & 60.99 & 53.21 & 60.71 & 51.41 \\
\midrule
\rowcolor{gray!15} \textbf{PMR (Ours)} & \textbf{76.76} & \textbf{69.45} & \textbf{87.60} & \textbf{70.70} & \textbf{33.59} & \textbf{30.24} & \textbf{25.86} & \textbf{25.14} & \textbf{63.82} & \textbf{53.98} & \textbf{61.70} & \textbf{56.56} \\
\bottomrule
\end{tabular}%
}
\caption{\textbf{Overall comparison across twelve evaluation settings.} PersonaMem (32K/128K) is in-domain; PrefEval (Explicit/Implicit), PersonaBench (4 noise levels), and PERMA (C-S/C-M/N-S/N-M denote Clean/Noise $\times$ Single/Multi-Domain) are out-of-domain. Higher is better. Best results are in \textbf{bold}.}
\label{tab:main}
\end{table*}

\paragraph{\textbf{Implementation.}}
We employ Qwen2.5-7B-Instruct~\cite{qwen} as the backbone language model for most methods, whereas Mem-$\alpha$ utilizes Qwen3-4B and MemSkill uses Qwen3-Next-80B-A3B-Instruct. For retrieval, all-MiniLM-L6-v2~\cite{minilm} extracts the top ten candidates. Our training set consists of 423 instances sampled from 70\% of the PersonaMem 32K subset. To reduce computational overhead, training uses retrieved dialogues as context, whereas inference utilizes the full history, processing context in 4K-token chunks. PMR runs 8 co-evolution rounds, where each round performs a full training pass over all training samples with $G{=}8$ rollout samples per query (temperature 1.0). The rubric reflection LLM is Qwen2.5-7B-Instruct, the same as the backbone, ensuring full self-containment. Key hyperparameters: credit modulation strength $\lambda{=}0.5$, endorsement temperature $\beta{=}2.0$, clip bound $\epsilon{=}0.28$, failure density gate $\theta_{\text{active}}{=}0.3$, cross-persona sharing threshold $\theta_{\text{flow}}{=}0.4$. Baselines are implemented using their public repositories; for a fair comparison, MemAgent, Mem-$\alpha$, and MemSkill are initialized from public checkpoints and fine-tuned on our 423 samples. All experiments run on 8 $\times$ NVIDIA A100 GPUs.

\subsection{Main Results}

\paragraph{\textbf{Overall Comparison with Baselines.}}
As shown in Tab.~\ref{tab:main}, PMR consistently demonstrates superior
performance across all twelve evaluation settings. On PersonaMem 32K, PMR
achieves \textbf{76.76} accuracy, outperforming the strongest baseline
MemCoE (67.47) by \textbf{9.29} points and the RL-based MemAgent (61.44)
by 15.32 points. This gap widens at the longer 128K context, where PMR
reaches 69.45 compared to MemCoE's 58.67, a \textbf{10.78}-point margin,
indicating that \emph{per-token credit assignment becomes increasingly
valuable as the number of sequential updates grows}. Static memory
pipelines (Mem0, A-Mem, LightMem) and context-based methods (LongContext,
RAG) trail significantly, confirming that learning from interaction
feedback is essential for long-horizon memory quality. Among RL-based
agents, MemSkill and MemCoE are the most competitive, yet they still lag
behind PMR by 10+ points on in-domain tasks. This advantage stems from
PMR's rubric-guided credit assignment, which provides \textbf{per-token diagnostic supervision} that neither scalar intermediate rewards (MemCoE) nor terminal trajectory-level rewards (MemAgent, Mem-$\alpha$, MemSkill) can offer.

\paragraph{\textbf{Generalization Across Settings.}}
PMR maintains strong performance on out-of-domain benchmarks. On PrefEval, PMR achieves \textbf{87.60} (Explicit) and 70.70 (Implicit), surpassing MemCoE by 4.20 and 2.10 points, respectively. On PersonaBench, PMR leads across all four noise levels, with the largest margin at noise=0 (33.59 vs 31.71), suggesting that the rubric's diagnostic evaluation is robust even in clean settings where other methods converge. On PERMA, PMR achieves the best results on C-S (63.82), N-S (61.70), and N-M (56.56), with the N-M gain of \textbf{5.15} points over MemCoE being particularly notable, as \emph{multi-domain noise settings stress-test the rubric's ability to handle heterogeneous failure patterns}. This generalization is driven by $\mathcal{R}_{\text{uni}}$, which captures transferable evaluation criteria across users, while the per-user layer's effect is \textbf{amortized into shared policy weights rather than stored as explicit per-user rules at inference}.

\begin{table}[t]
\centering
\small
\colorlet{dropA}{red!6}
\colorlet{dropB}{red!15}
\colorlet{dropC}{red!24}
\colorlet{dropD}{red!33}
\colorlet{dropE}{red!42}
\setlength{\tabcolsep}{3.1pt}
\begin{tabular}{@{}l cc cccc@{}}
\toprule
& \multicolumn{2}{c}{\textbf{PersonaMem}} & \multicolumn{4}{c}{\textbf{PERMA}} \\
\cmidrule(lr){2-3} \cmidrule(lr){4-7}
\textbf{Configuration} & 32K & 128K & C-S & C-M & N-S & N-M \\
\midrule
\rowcolor{gray!15} \textbf{PMR (Full)} & \textbf{76.76} & \textbf{69.45} & \textbf{63.82} & \textbf{53.98} & \textbf{61.70} & \textbf{56.56} \\
w/o Positive-only            & \cellcolor{dropE}57.23 & \cellcolor{dropE}51.34 & \cellcolor{dropE}46.24 & \cellcolor{dropD}40.36 & \cellcolor{dropD}47.52 & \cellcolor{dropE}39.07 \\
w/o Rubric amplitude ($w_t$) & \cellcolor{dropD}62.16 & \cellcolor{dropE}52.81 & \cellcolor{dropE}47.52 & \cellcolor{dropE}35.99 & \cellcolor{dropE}46.24 & \cellcolor{dropD}41.90 \\
w/o Diagnostic (5-dim)       & \cellcolor{dropD}63.85 & \cellcolor{dropD}58.05 & \cellcolor{dropD}51.06 & \cellcolor{dropC}46.27 & \cellcolor{dropD}50.35 & \cellcolor{dropC}49.36 \\
w/o Entropy gate             & \cellcolor{dropD}65.66 & \cellcolor{dropC}60.24 & \cellcolor{dropB}59.57 & \cellcolor{dropA}52.18 & \cellcolor{dropB}57.16 & \cellcolor{dropB}51.41 \\
w/o Direction gate           & \cellcolor{dropC}67.03 & \cellcolor{dropC}61.61 & \cellcolor{dropC}57.02 & \cellcolor{dropB}49.61 & \cellcolor{dropB}58.15 & \cellcolor{dropA}53.98 \\
\bottomrule
\end{tabular}
\caption{Ablation on credit assignment components. Cell color indicates drop from Full (\textcolor{dropA}{\rule{0.8em}{0.8em}}~$<$3 to \textcolor{dropE}{\rule{0.8em}{0.8em}}~$\geq$15\,pp). Rows ordered by average drop.}
\vspace{-0.8cm}
\label{tab:ablation-credit}
\end{table}

\subsection{Ablation Studies}
\label{sec:ablation}

We conduct ablation experiments to isolate the contribution of each design choice in PMR. All ablations are evaluated on PersonaMem (32K/128K) and PERMA (C-S/C-M/N-S/N-M), following the same setting as the main experiment.

\paragraph{\textbf{Credit Assignment Components.}}
Tab.~\ref{tab:ablation-credit} isolates each credit assignment component by removing it from the full model. Removing the rubric amplitude ($w_t$) causes the largest drop on PersonaMem (\textbf{14.6} and \textbf{16.6} points at 32K and 128K), confirming that per-token credit modulation based on rubric endorsement is the core mechanism; without it, all tokens receive uniform trajectory-level advantage, collapsing to pure GRPO. The direction gate, which discards endorsements that disagree with the trajectory outcome instead of letting them shrink a token's credit, contributes up to 9.7 points across settings. The entropy gate shows the widest variation: \textbf{11.1} points on PersonaMem 32K, where long histories produce many carry-over tokens to suppress, but only 1.8 on PERMA C-M, where moderate noise leaves fewer carry-overs. Removing the positive-only restriction produces the largest drop on PERMA (up to \textbf{17.6} points on C-S), as applying rubric-guided credit to failed trajectories introduces unreliable blame attribution. Finally, removing the five-dimensional diagnostic rubric (reverting to scalar endorsement) costs \textbf{12.9} and \textbf{11.4} points on PersonaMem, confirming that \emph{per-dimension diagnosis carries richer signal than a single score}. The two largest contributions target different regimes: $w_t$ dominates on PersonaMem where long histories produce many edit decisions to sort, while positive-only dominates on PERMA where high noise makes failed-trajectory blame attribution especially fragile, suggesting that credit modulation and blame avoidance address complementary failure modes. Together, the three signal components ($w_t$, positive-only, diagnostic rubric) account for the bulk of PMR's gains, while the two gates provide complementary safety benefits at lower individual cost.  

\paragraph{\textbf{Rubric Structure.}}
To isolate the contribution of each rubric structural component, we remove one at a time from the full model (Tab.~\ref{tab:ablation-rubric}). The most striking finding is that \textbf{disabling cross-persona sharing causes the largest degradation across all settings}, exceeding the removal of either rubric layer individually. This indicates that the \emph{dynamic boundary} between $\mathcal{R}_{\text{uni}}$ and $\mathcal{R}_{\text{spec}}$ is \textbf{more critical than the content of either layer}: per-user discoveries must flow upward to benefit other users, and blocking this transfer hurts more than losing per-user adaptation entirely. The \textbf{coverage test ranks second}, confirming that forcing per-user rules when $\mathcal{R}_{\text{uni}}$ suffices introduces substantial noise through \emph{over-personalization}. Removing $\mathcal{R}_{\text{uni}}$ hurts more than removing $\mathcal{R}_{\text{spec}}$, consistent with the universal layer serving as the \textbf{shared foundation}. Notably, $\mathcal{R}_{\text{uni}}$ removal causes \textbf{larger drops on out-of-domain benchmarks than in-domain}, suggesting that the universal layer's transferability is \textbf{key to generalization}; $\mathcal{R}_{\text{spec}}$ removal shows the opposite pattern, as per-user adaptation is less relevant for unseen users whose per-user layer is amortized into weights rather than stored explicitly. The quality gate has a \textbf{moderate effect}, indicating that non-discriminative rubrics introduce meaningful but not catastrophic noise.

\begin{figure}[t]
\centering
\includegraphics[width=\columnwidth]{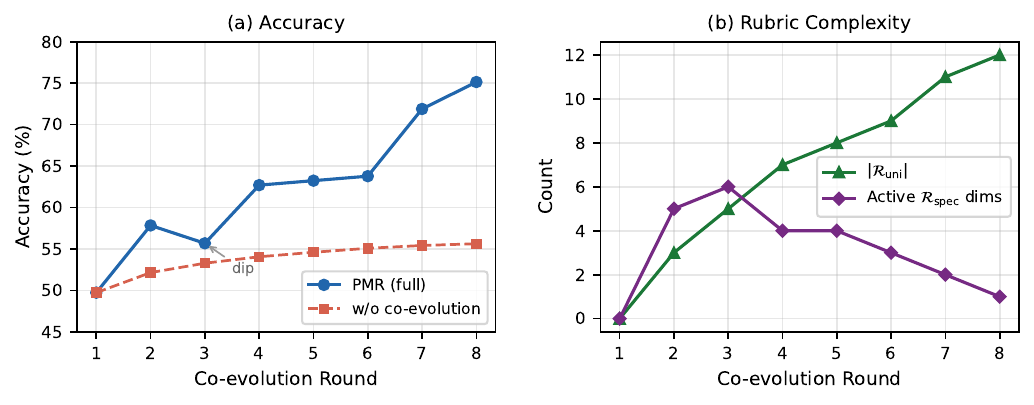}
\caption{Co-evolution dynamics across 8 rounds on PersonaMem 32K. Left axis: accuracy; PMR (full, rising) and w/o co-evolution (frozen rubric, plateauing). Right axis: $|\mathcal{R}_{\text{uni}}|$ (growing) and active $\mathcal{R}_{\text{spec}}$ (decreasing as student learns).}
\label{fig:coevolution}
\vspace{-0.2cm}
\end{figure}

\paragraph{\textbf{Co-Evolution Dynamics.}}
Fig.~\ref{fig:coevolution} traces accuracy and rubric complexity across co-evolution rounds. Starting from an \emph{empty rubric} in round~1, the universal layer acquires checks monotonically as cross-persona failure patterns are discovered, while the per-user layer's active dimensions \textbf{first grow and then transition to resolved} as the student learns to address user-specific failure modes. A \textbf{temporary accuracy dip at round~3} coincides with a spike in active $\mathcal{R}_{\text{spec}}$ dimensions: newly induced checks temporarily destabilize the student, but \textbf{accuracy recovers by round~4} as distillation absorbs the tightened rubric. In contrast, the \emph{w/o co-evolution} baseline, which freezes the rubric after the initial induction, improves only marginally and converges to a \textbf{substantially lower accuracy}. This contrast confirms that \textbf{rubric refinement must track the student's evolving failure landscape}.
\begin{table}[t]
\centering
\small
\colorlet{dropA}{red!6}
\colorlet{dropB}{red!15}
\colorlet{dropC}{red!24}
\colorlet{dropD}{red!33}
\colorlet{dropE}{red!42}
\setlength{\tabcolsep}{3.1pt}
\begin{tabular}{@{}l cc cccc@{}}
\toprule
& \multicolumn{2}{c}{\textbf{PersonaMem}} & \multicolumn{4}{c}{\textbf{PERMA}} \\
\cmidrule(lr){2-3} \cmidrule(lr){4-7}
\textbf{Configuration} & 32K & 128K & C-S & C-M & N-S & N-M \\
\midrule
\rowcolor{gray!15} \textbf{PMR (Full)} & \textbf{76.76} & \textbf{69.45} & \textbf{63.82} & \textbf{53.98} & \textbf{61.70} & \textbf{56.56} \\
w/o Cross-persona sharing & \cellcolor{dropE}54.05 & \cellcolor{dropE}45.76 & \cellcolor{dropE}42.69 & \cellcolor{dropE}34.70 & \cellcolor{dropE}42.98 & \cellcolor{dropE}34.45 \\
w/o Coverage test           & \cellcolor{dropE}59.63 & \cellcolor{dropE}52.99 & \cellcolor{dropE}46.10 & \cellcolor{dropD}38.30 & \cellcolor{dropE}42.98 & \cellcolor{dropE}35.99 \\
w/o $\mathcal{R}_{\text{uni}}$ (per-user only) & \cellcolor{dropD}66.49 & \cellcolor{dropC}62.96 & \cellcolor{dropD}51.06 & \cellcolor{dropD}41.90 & \cellcolor{dropD}49.22 & \cellcolor{dropD}44.21 \\
w/o Quality gate            & \cellcolor{dropC}67.03 & \cellcolor{dropC}62.56 & \cellcolor{dropC}56.45 & \cellcolor{dropC}46.27 & \cellcolor{dropC}54.33 & \cellcolor{dropC}46.79 \\
w/o $\mathcal{R}_{\text{spec}}$ (global only) & \cellcolor{dropC}70.27 & \cellcolor{dropB}66.01 & \cellcolor{dropB}58.15 & \cellcolor{dropA}50.38 & \cellcolor{dropA}58.72 & \cellcolor{dropB}51.41 \\
\bottomrule
\end{tabular}
\caption{Ablation on rubric structure components. Cell color indicates drop from Full (\textcolor{dropA}{\rule{0.8em}{0.8em}}~$<$3 to \textcolor{dropE}{\rule{0.8em}{0.8em}}~$\geq$15\,pp).}
\vspace{-1cm}
\label{tab:ablation-rubric}
\end{table}

\begin{figure*}[t]
    \centering
    \includegraphics[width=\linewidth]{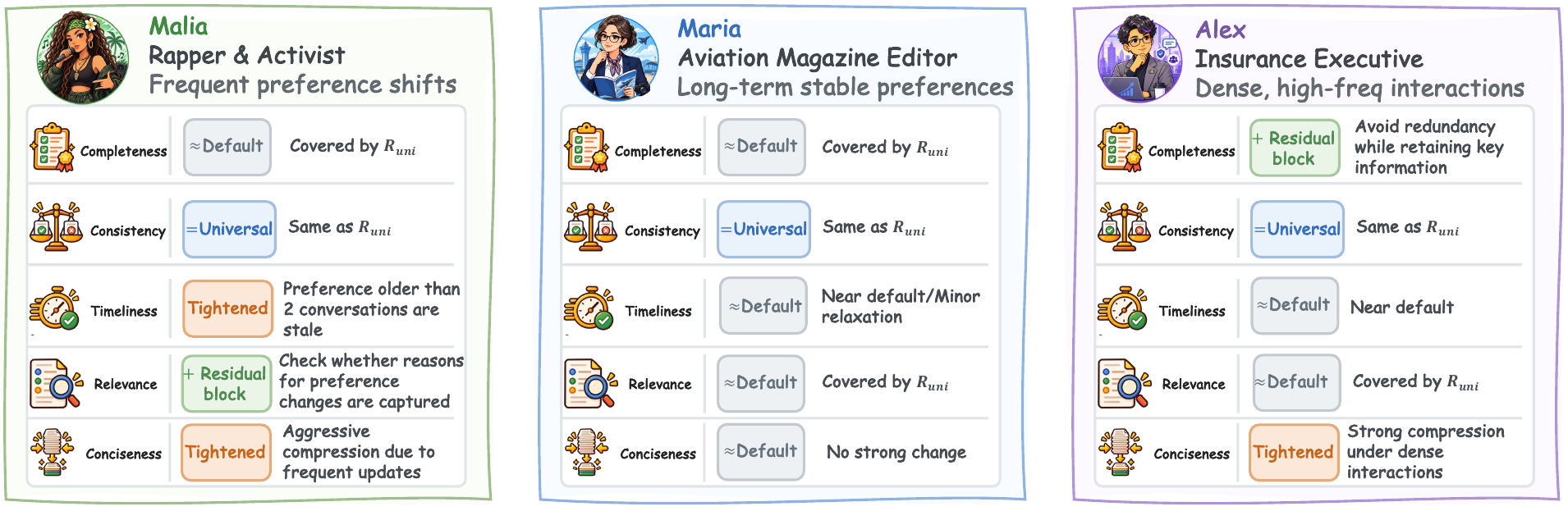}
    \caption{Case study of learned personalized rubrics. PMR selectively adapts rubric dimensions to user behavior: Malia emphasizes timeliness and conciseness, Maria largely retains universal criteria, and Alex strengthens completeness and conciseness. Consistency remains universal across all three personas.}
    \label{fig:rubric-case}
\end{figure*}

\subsection{Further Analysis}

\begin{figure}
    \centering
    \includegraphics[width=\linewidth]{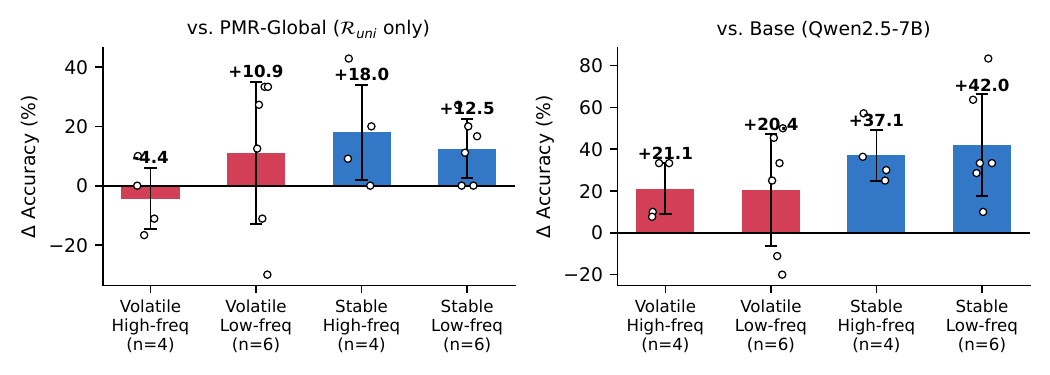}
    \caption{Per-user gains from $\mathcal{R}_{\text{spec}}$ across four persona groups defined by median splits of volatility and interaction frequency. $\Delta = \text{Acc}(\text{PMR}) - \text{Acc}(\text{PMR-Global})$, with larger gains for stable users.}
    \label{fig:per-user}
    \vspace{-0.6cm}
\end{figure}

\paragraph{\textbf{Per-User Analysis.}}
To examine how personalization benefits different users, we partition the 20 PersonaMem personas into four groups based on median splits of preference volatility and interaction frequency. Fig.~\ref{fig:per-user} reports the accuracy gain of PMR over PMR-Global ($\mathcal{R}_{\text{uni}}$ only). Surprisingly, stable users benefit more from $\mathcal{R}_{\text{spec}}$ than volatile users. One possible explanation is that volatile users share common failure modes, such as stale preferences, that the universal rubric already addresses, whereas stable users exhibit more idiosyncratic recall and consistency errors requiring user-specific criteria. This suggests that personalization adds the most value where universal guidance falls short, rather than where preferences change most frequently. The volatile, high-frequency group is an exception ($-4.4$\,pp), potentially because preferences evolve faster than reliable user-specific evidence accumulates, a limitation only partially mitigated by the quality gate.

\paragraph{\textbf{Rubric Interpretability}}

Fig.~\ref{fig:rubric-case} illustrates the learned personalized rubrics for three representative personas across five dimensions.
For \textbf{Malia}, whose preferences shift frequently, $\mathcal{R}_{\text{spec}}$ tightens \textit{Timeliness} by treating preferences older than two conversations as stale, strengthens \textit{Conciseness} for frequent updates, and adds a \textit{Relevance} check to capture reasons for preference changes. For \textbf{Maria}, whose preferences remain stable, the rubric largely retains universal criteria, with only minor adjustments to \textit{Timeliness}. For \textbf{Alex}, whose interactions are frequent and dense, $\mathcal{R}_{\text{spec}}$ adds a \textit{Completeness} check to reduce redundancy while preserving key information and strengthens \textit{Conciseness} for greater compression. Notably, \textit{Consistency} remains universal across all three personas. Overall, these cases demonstrate \textbf{selective, behavior-aligned specialization}: personalized rubrics adapt to user-specific needs while preserving universal criteria elsewhere.

\begin{figure}
  \centering
  \includegraphics[width=\linewidth]{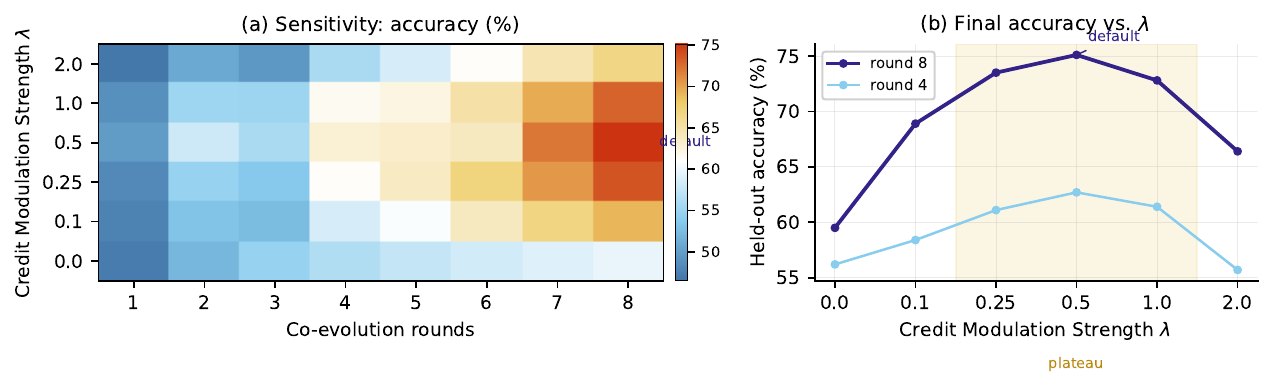}
  \caption{Sensitivity to credit modulation strength $\lambda$ and co-evolution rounds. (a) Held-out accuracy across the $\lambda \times$ rounds grid. (b) Accuracy versus $\lambda$ at rounds~4 and~8, with shading indicating the robust plateau ($\lambda \in [0.25,1.0]$).}
  \label{fig:sensitivity}
  \vspace{-0.4cm}
\end{figure}

\paragraph{\textbf{Sensitivity to $\lambda$ and Co-evolution Rounds.}}
Fig.~\ref{fig:sensitivity} examines the interplay between credit modulation strength \(\lambda\) and co-evolution rounds, revealing three key findings. \textbf{(i) Rubric modulation enables sustained improvement.} Without modulation (\(\lambda{=}0\)), training saturates early, whereas \(\lambda{=}0.5\) continues to yield substantial gains in later rounds. This highlights the importance of rubric-guided credit differentiation beyond outcome-only supervision. \textbf{(ii) Performance is robust to \(\lambda\).} Accuracy peaks at the default \(\lambda{=}0.5\) but remains within \(2.3\) pp across \(\lambda\in[0.25,1.0]\), indicating a broad, stable optimum rather than sensitivity to precise tuning. \textbf{(iii) Benefits compound with co-evolution.} The advantage over standard GRPO grows from \(2.7\) pp to \(15.6\) pp across rounds, suggesting that the co-evolving rubric provides increasingly informative credit signals. Even aggressive modulation (\(\lambda{=}2.0\)) outperforms the unmodulated baseline despite transient fluctuations. The direction gate and bounded credit adjustment (Sec.~\ref{sec:method-credit}) help explain this graceful degradation under excessive modulation.

\section{Conclusion}
\label{sec:conclusion}

In this paper, we introduced \textbf{Personalized Memory Rubrics} (PMR), a framework that transforms sparse terminal QA feedback into fine-grained, per-user credit assignment for long-horizon memory agents. PMR learns a diagnostic rubric decomposed into universal and user-adaptive layers from success/failure rollout comparison, and uses it as privileged context for a frozen teacher to redistribute credit at the token level within GRPO. The design rests on three key principles: the rubric modulates only credit \emph{magnitude} while direction is always determined by real QA outcomes (structural anti-circularity); the asymmetric positive-only design exploits the reliability gap between credit sorting on successful trajectories and blame attribution on failed ones; and the co-evolution loop allows the rubric to track the student's evolving failure landscape. Unlike prior methods that depend on an external LLM judge at inference, PMR is fully self-contained. Experiments across four benchmarks demonstrate consistent improvements, with ablations confirming that each design choice contributes to performance.


\section*{Ethical Considerations}
All experiments in this work are performed using publicly accessible benchmark datasets that have been fully anonymized and contain no personally identifiable or sensitive information. The study does not involve human participants, protected demographic attributes, or data that may pose privacy risks. We do not anticipate any significant negative societal consequences, including discriminatory outcomes, unauthorized surveillance, or intentional misuse by malicious actors. The proposed PMR framework is intended solely for research on personalized agent memory and is not developed for deployment in sensitive or safety-critical scenarios. Throughout the study, we adhere to standard ethical practices concerning data use, reproducibility, and responsible research. Experimental configurations and implementation details are reported to promote transparency and facilitate replication.

\bibliographystyle{ACM-Reference-Format}
\bibliography{9Reference}

\end{document}